\pdfoutput=1
\documentclass[11pt]{article}

\usepackage[final]{acl}

\usepackage{times}
\usepackage{latexsym}
\usepackage{amsmath} 
\usepackage{booktabs}
\usepackage[T1]{fontenc}

\usepackage{xcolor}
\usepackage{colortbl}
\usepackage[utf8]{inputenc}
\usepackage{subfig}
\usepackage{graphicx}
\usepackage{microtype}
\usepackage{multirow}
\usepackage{inconsolata}

\usepackage{graphicx}

\definecolor{lightLavender}{HTML}{EEEBFA}
\title{Sparse-to-Dense: A Free Lunch for Lossless Acceleration of Video Understanding in LLMs}

\author{
 \textbf{Xuan Zhang\textsuperscript{1}},\quad
  {\textbf{Cunxiao Du}}\textsuperscript{2}\thanks{Corresponding author: \texttt{ducx@sea.com}},
  \quad
 \textbf{Sicheng Yu\textsuperscript{1}},\quad
 \textbf{Jiawei Wu\textsuperscript{3}},\\
 \textbf{Fengzhuo Zhang\textsuperscript{3}},\quad
 \textbf{Wei Gao\textsuperscript{1}},\quad
 \textbf{Qian Liu\textsuperscript{2}}
\\
 \textsuperscript{1}Singapore Management University,\quad
 \textsuperscript{2}Sea AI Lab,\quad
 \textsuperscript{3}National University of Singapore\\
}

\begin{document}
\maketitle
\begin{abstract}
% Recent advancements have integrated large language models (LLMs) with video understanding capabilities, demonstrating exceptional performance across a wide range of tasks. 
Video Large Language Models (Video-LLMs) suffer from high inference latency in long video processing due to their auto-regressive decoding mechanism, posing challenges for the efficient processing of video sequences that are usually very long. 
We observe that attention scores in Video-LLMs during decoding exhibit pronounced sparsity, with computational focus concentrated on a small subset of critical tokens.
Motivated by this insight, we introduce Sparse-to-Dense (\textsc{StD}), a novel decoding strategy that integrates two distinct modules: a sparse module that rapidly generates speculative tokens using efficient top-$K$ attention, and a dense module that verifies these tokens in parallel via full self-attention.
This collaborative approach accelerates Video-LLMs losslessly, effectively offering a free lunch for video understanding.
\textsc{StD} is a plug-and-play solution requiring no fine-tuning or architectural changes and achieves up to a 1.94$\times$ wall time speedup while preserving model performance. It enables a seamless conversion of standard Video-LLMs into sparse counterparts, unlocking efficient long-video processing without sacrificing accuracy.
% \qian{Video Large Language Models (Video-LLMs) suffer from high inference latency in long video processing due to their auto-regressive decoding mechanism.
% We observe that attention scores in Video-LLMs during decoding exhibit pronounced sparsity, with computational focus concentrated on a small subset of critical tokens.
% Motivated by this insight, we introduce Sparse-to-Dense (\textsc{StD}), a novel decoding strategy that integrates two distinct modules: a sparse module that rapidly generates speculative tokens using efficient top-$K$ attention, and a dense module that verifies these tokens in parallel via full self-attention.
% This collaborative approach accelerates Video-LLMs losslessly, effectively offering a free lunch for video understanding.
% \textsc{StD} is a plug-and-play solution requiring no fine-tuning or architectural changes, and achieves up to a 1.94$\times$ walltime speedup while preserving model performance. It enables a seamless conversion of standard Video-LLMs into sparse counterparts, unlocking efficient long-video processing without sacrificing accuracy.
% }
\end{abstract} 
\section{Introduction}
\label{sec:intro}

% Recent studies have integrated large language models (LLMs) with video understanding capabilities, i.e., Video Large Language Models (Video-LLMs), and demonstrated exceptional performance across a wide range of tasks, such as video question answering and video captioning~\cite{lin2024vila,cao2024madtp,zhang2025videollama}.
% Researchers have pushed the input frames to about 6K frames~\citep{zhang2024long,xue2024longvila}.
% A common practice in Video-LLMs is representing a video as a sequence of image frames, resulting in long token sequences. For instance, processing a 1-hour video by extracting one frame every $5$ seconds results in $720$ frames, which is approximately $141,120$ visual tokens in VILA~\cite{lin2024vila}.
Recent advances in Video Large Language Models (Video-LLMs), which combine large language models with video understanding, have achieved exceptional performance on tasks like video question answering and captioning~\cite{lin2024vila,cao2024madtp,zhang2025videollama}. A common practice in Video-LLMs is representing a video as a sequence of image frames, which results in extremely long token sequences that can strain computational resources. 
For instance, a 1-hour video sampled at 5-second intervals produces 720 frames, which translates to 141,120 visual tokens in VILA~\cite{lin2024vila}. These extremely long token sequences cause Video-LLMs to suffer from high inference latency when processing lengthy videos, making real-time applications challenging.

% \qian{Recent advances in Video Large Language Models (Video-LLMs), which combine large language models with video understanding, have achieved exceptional performance on tasks like video question answering and captioning~\cite{lin2024vila,cao2024madtp,zhang2025videollama}. A common practice in Video-LLMs is representing a video as a sequence of image frames, which results in extremely long token sequences that can strain computational resources. For instance, a 1-hour video sampled at 5-second intervals produces 720 frames, which translates to \~141,120 visual tokens in VILA~\cite{lin2024vila}. These extremely long token sequences cause Video-LLMs to suffer from high inference latency when processing lengthy videos, making real-time applications challenging.}

%Due to the auto-regressive nature of current Video-LLMs, each new token is generated based on all the preceding tokens, which creates substantial I/O and computation challenges for inference.
This latency is primarily introduced by the auto-regressive nature of current Video-LLMs, where each new token must attend to all preceding tokens, creating substantial memory and computational challenges.% during inference.
% \qian{This latency is primarily introduced by the auto-regressive nature of current Video-LLMs, where each new token must attend to all preceding tokens, creating substantial memory and computational challenges during inference.}
%when processing videos represented as long token sequences.
While mechanisms like key-value (KV) caching are employed to store pre-computed key and value tensors and reduce redundant re-computation, frequent access to the cache imposes heavy demands on memory bandwidth due to the growing amount of KV cache with the increasing sequence length.
%\qian{Try to avoid the usage of I/O snce it is too professional here. Maybe use memory would be better.}
%The amount of KV cache, which is $O(\text{seq\_len} \times \text{head\_num} \times \text{head\_dim})$, grows proportionally with the sequence length.
%\qian{Polish this, can just say that KV cache grows with the sequence length?}
%The KV cache occupies a large amount of VRAM and causes repeated memory operations on the cache during each decoding step. 
This significantly reduces the throughput of Video-LLMs. A common approach to addressing this problem is KV cache compression~\cite{du2024revisiting, chen2024image,lin2024boosting,zhang2025lighttransfer} or quantization~\cite{su2025akvq,hooper2024kvquant,liu2024kivi} at test time. However, these methods introduce discrepancies between training and inference, degrading the performance of LLMs.
In this paper, we aim to build a lossless acceleration method designed specifically for Video-LLMs that preserves the exact output distribution of the original model. Although speculative decoding~\cite{leviathan2023fast,chen2023accelerating,hou2025banditspec} meets this requirement, it usually requires an extra draft model, which is expensive for Video-LLMs. In contrast, we observe that Video-LLMs exhibit a unique structural property, attention sparsity, which can serve as a training-free and plug-and-play draft model. Specifically, retaining only the top-$K$ KV caches in the attention layers preserves the original predictions for approximately 95\% of tokens (empirically verified), suggesting that most attention heads contribute minimally to the final output.
%\qian{In this paper, we propose a lossless acceleration method designed specifically for Video-LLMs that preserves the exact output distribution of the original model. While traditional speculative decoding techniques~\cite{leviathan2023fast,chen2023accelerating,xia2023speculative} achieve lossless acceleration for standard LLMs by leveraging a draft model, this approach becomes impractical in the context of Video-LLMs due to the inherent computational complexity of Video-LLMs. However, we observe that Video-LLMs exhibit a unique structural property, attention sparsity. Specifically, retaining only the top-$K$ KV caches in the attention layers preserves the original predictions for approximately 95\% of tokens (empirically verified), suggesting that most attention heads contribute minimally to the final output.}
Motivated by this observation, we introduce a novel decoding method called Sparse-to-Dense (\textsc{StD}), which leverages the sparse structure of Video-LLMs as its draft model. This design eliminates the need for an extra trained draft model, making \textsc{StD} a plug-and-play solution.
We refer to the original Video-LLM as the dense model because it decodes using the full KV cache, whereas the model with top-$K$ attention is termed the sparse model.
Both models share identical architectures, differing only in how they compute attention.
Therefore, we do not need additional GPU memory to store the sparse model, nor does it require any extra training. 
The top‑$K$ attention in the sparse model boosts decoding speed while sacrificing some token quality, whereas the dense model is slower but guarantees accuracy.
% Inspired by the recent successes of speculative decoding, we employ a model with sparse attention as a draft model to auto-regressively propose the next $\gamma$ tokens, while using the original model with dense attention as a verifier to conduct parallel verifications to avoid multiple times I/O of full KV cache. 
% This setup guarantees that the outputs from our sparse-to-dense framework match those of the original Video-LLM without any discrepancy.
We use the sparse model to auto-regressively draft the next \(\gamma\) tokens, while the dense model verifies them in parallel. This approach avoids redundant full KV cache memory and ensures the outputs exactly match those of the original Video-LLM.

We conduct experiments on representative Video-LLMs including LLaVA-OneVision~\cite{li2024llava} and Qwen2-VL~\cite{wang2024qwen2}, evaluating them
on video understanding benchmarks like MLVU~\cite{zhou2024mlvu} and VideoMME~\cite{fu2024video}.
Experiment results show that our \textsc{StD}, serving as a tuning-free, plug-and-play solution, achieves up to a 1.94$\times$ of video input processing without any performance degradation. It is immediately deployable, requiring only 20 lines of code to transform an original Video-LLM into a sparse Video-LLM, and it does not require any extra training to deploy the draft model.

\section{Observation}
\label{sec:obs}

In this section, we investigate the disparity in decoded tokens between two configurations of Video-LLMs: 1) sparse top-$K$ KV cache: utilizing only the top-$K$ KV caches based on the highest attention weights; and 2) dense full KV cache: employing the complete set of KV caches.
We conduct experiments using the Qwen2-VL-7B~\cite{wang2024qwen2} model on randomly selected samples from MLVU~\cite{zhou2024mlvu}, and Video-MME~\cite{fu2024video} datasets. 
We evaluate the next-token prediction accuracy of the model when employing sparse attention with top-$K$ KV caches.
Our findings indicate that the model with sparse attention maintains an average token prediction accuracy exceeding 95\%. 
This high accuracy suggests that for the majority of decoded tokens, only the top-$K$ KV caches are necessary. 
However, it is important to note that the 95\% accuracy is measured per individual token and does not accumulate across multiple tokens. 
For instance, the accuracy of correctly predicting five consecutive tokens drops to approximately $(95\%)^5 \approx 77\%$.

% Therefore, while using only the top-$k$ attention significantly reduces computational and memory overhead, it does not guarantee lossless decoding in all scenarios, especially when multiple tokens need to be predicted accurately. 
% Nonetheless, for most tokens, the top-$k$ KV caches provide an effective balance between efficiency and performance.

% In this section, we analyze the gap of decoded tokens between 1) a video-LLM using sparse top-$k$ KV caches (i.e., those allocated with the highest attention weights) and 2) a video-LLM using a dense full KV cache.
% This observation provides insights that motivate our approach to losslessly accelerate the inference time of video-LLM based on the collaborative work of the same model with both sparse and dense attention.
% To investigate this, we conduct experiments on MLVU, and Video-MME (with short, medium, and long videos respectively) using Qwen2-VL-7B.
% We compute the next token prediction accuracy of the model with sparse attention, and find that on average accuracy is more than 95.x\%, which suggest that for most of the decoded tokens, only top-$k$ KV caches are needed, and for a small portion of decoding tokens, need full attention. 
% Note that 95.x\% is not a accumulated value, it's only the accuracy for one token. 
% In other words, the accuracy of predicting correct 5 tokens is about 77.4\%.
% Therefore, it's not lossless to directly use top $k$ attention when decoding, and the worst case can not be guaranteed, although for most tokens, only top $k$ attention is  enough.

\section{Method}
\label{sec:method}
In this section, we present Sparse-to-Dense (\textsc{StD}), a method designed to achieve lossless acceleration for Video-LLMs. 
We refer to the original model $\mathcal{M}$ as the dense model, as it requires the full KV cache during decoding, while the sparse model $\mathcal{M}_s$ uses sparse attention. 
Although $\mathcal{M}_s$ is faster, it is somewhat less accurate.
Unlike traditional speculative decoding, which relies on an additional draft model, our approach leverages \(\mathcal{M}_s\) with the same parameters as \(\mathcal{M}\). The only difference is that \(\mathcal{M}_s\) loads a reduced KV cache to perform sparse attention, eliminating the need for extra GPU memory to store another model's parameters.
In the following subsections, we will detail the decoding procedure and the design of the sparse model.
% overview, 有一个model_s, model_d, 然后model_s 推n个词, model_d with causal mask 并行验证.
% refer一下是无损的, 点名attention的复杂度
\subsection{Decoding Procedures}
In our \textsc{StD}, the sparse model $\mathcal{M}_s$ functions as a draft model to propose potential next $\gamma$ tokens, while the dense model $\mathcal{M}$ verifies them to derive the final output sequence. 
Given an input sequence $\{x_0, \cdots, x_{m-1}\}$, consisting of visual %$X_v$ 
and textual tokens 
%$X_t$
, the sparse model $\mathcal{M}_s$ auto-regressively generates $\gamma$ subsequent token candidates $\{x_m, \cdots, x_{m+\gamma-1}\}$.
Because the tokens proposed by the sparse model $\mathcal{M}_s$ might not align with those predicted by the dense model $\mathcal{M}$, it requires the verification of $\mathcal{M}$.
The dense model $\mathcal{M}$ verifies all $\gamma$ proposed tokens in parallel, requiring only a single I/O operation for the full KV cache.
Thus, this verification procedure accelerates the process compared with the auto-regressive decoding of $\mathcal{M}$ itself, where each token requires a separate I/O operation.
During the verification, $\mathcal{M}$ identifies the first $n$ tokens that align with its predictions, where $0 \leq n \leq \gamma$, and additionally provides a bonus token $\hat{x}_{n+m}$ for free. 
The verified sequence $\{x_m, \cdots, x_{m+n-1}, \hat{x}_{n+m}\}$ is then appended to the input sequence $\{x_0, \cdots, x_{m-1}\}$ to form the context for the next round of proposal and verification.
%There are two acceptance strategies: greedy decoding and rejection sampling, both applicable to our method. 
%These strategies correspond to the outputs of the dense model $\mathcal{M}$ when using greedy search and random sampling decoding strategies, respectively.

% 推gamma个token，m_s是gamma*（k+m_t)， 并行验证是（m_v+m_t), 相加后/n+1，
% 原来是n+1 *（m_v+m_t)

 % 算出理论上对KV cache的I/O复杂度
 
\subsection{Model with Sparse Attention}
%\subsection{Model $\mathcal{M}_s$ with Sparse Attention}
Next, we introduce the design of our sparse model $\mathcal{M}_s$.
Empirical observations in Section~\ref{sec:obs} 
indicate that during most decoding steps, attention scores are predominantly concentrated on a small subset of KV caches, a pattern we term \textit{sparse} attention (also known as top-$K$ attention~\cite{lou2024sparser}). 
Only a small fraction of tokens require more evenly distributed \textit{dense} attention. 
This insight motivates a strategy to selectively apply sparse attention for the majority of tokens and resort to dense attention only when necessary, reducing the I/O overhead of accessing the full KV cache %compared with always using full attention
, and thereby improving decoding speed.

Since the number of visual tokens is typically much larger than the number of textual tokens ($m_v \gg m_t$), with $m_v$ often exceeding 10,000 while $m_t$ are usually around 100, our primary focus is on reducing the size of the visual KV cache. 
To achieve this, we leverage the attention patterns of the textual tokens $X_t$ to identify and select the most relevant KV caches from the visual tokens.
Specifically, we analyze the allocation of attention scores when processing the textual tokens $X_t = \{x_{m_v}, \cdots, x_{m-1}\}$ (i.e., the last $m_t$ tokens in the input sequence) to identify which KV pairs of the visual tokens $X_v$ contribute more during the prefilling stage.
For each layer $l$, we calculate the average attention scores directed toward the visual tokens $X_v$ for textual tokens $X_t$.
We then retain only the top-$K$ KV pairs of visual tokens with the highest attention scores.
%Note that ideally, the top $k$ KV caches should be re-selected at each decoding step like Quest~\cite{}, as the attention focus may vary for each token. 
%However, dynamically finding the top $k$ KV caches during decoding can be time-consuming. 
To balance performance and efficiency, we determine the retained $K$ KV caches only during the prefilling stage and avoid the computation-demand dynamic selections in the decoding stage.
The selected visual tokens can vary across different layers and attention heads, reflecting the distinct focus of each layer and head in processing the input.
The selection of the KV cache of layer $l$ can be formalized as
\vspace{-0.8em}
\begin{align*}
\text{Cache}_s[l]= 
\text{argTopK}_{x \in X_v}\bigg(\frac{1}{m_t}\sum_{\hat{x} \in X_t} A_l(\hat{x}, x)\bigg),
\end{align*}

\vspace{-0.9em}
where $\text{argTopK}(\cdot)$ is an operation that selects the top-$K$ elements indices with the highest values from a given set, $k$ is a predefined hyper-parameter, and $A_l(\hat{x}, x)$ represents the attention score from token $\hat{x}$ to token $x$ in layer $l$.   
For models utilizing Grouped Query Attention (GQA)~\cite{ainslie2023gqa}, where the number of query heads equals the number of groups multiplied by the number of KV heads, we directly sum the attention scores within each group to select the top-$K$ KV caches for this head.
The KV cache selection operates at the granularity of individual KV heads, allowing each layer or head to retain a distinct subset of caches based on its specific requirements.
%Note that the architectures of the dense model $\mathcal{M}$ and the sparse model $\mathcal{M}_s$ are identical. The only difference lies in their attention mechanisms: the dense model $\mathcal{M}$ uses full attention, whereas the sparse model $\mathcal{M}_s$ operates only on the selected KV caches, $\text{Cache}_s[l]$.

% model_s的设计
% topk attention identification
% part 1, 拿所有的language作为query去选topk 选择
% part2, attn选择
% 可以贴一下pesudo code, 简单有效
% When decode the first $n+1$ token, when use our Sparse-to-Dense decoding, the overall I/O for input sequence during sparse model to speculatively propose is $\gamma*(k+m_t)$, for dense model to verify in parallel is $(m_v + m_t)$, so the average I/O is $\frac{\gamma*(k+m_t)+m_v + m_t}{n+1}$, while the original I/O using vanilla decoding is $(n+1)*(m_v + m_t)$.
\subsection{I/O complexity analysis.}

 In the decoding phase, the I/O complexity of our Sparse-to-Dense decoding method can be analyzed as follows. 
For the sparse model $\mathcal{M}_s$, which speculatively proposes $\gamma$ subsequent tokens, the I/O cost involves accessing the selected $K$ visual KV caches and all $m_t$ textual KV caches. Thus, the total I/O for the sparse model is given by: $\text{I/O}_{\text{sparse}} = \gamma \times (K + m_t)$. 
For the dense model $\mathcal{M}$, which verifies the proposed tokens in parallel, the I/O cost includes accessing the full KV caches of all visual and textual tokens, resulting in:
$\text{I/O}_{\text{dense}} = m_v + m_t$.
The total I/O for Sparse-to-Dense decoding is therefore:
$\text{I/O}_{\text{total}} = \gamma \times (K + m_t) + (m_v + m_t)$,
and the average I/O per token is %\qian{This seems overflowed, I have made this smaller using \small}
\vspace{-0.3em}
\begin{align*}
\small
    \text{I/O}_{\text{average}} = \frac{\text{I/O}_{\text{total}}}{\alpha\times \gamma} = \frac{\gamma \times (K + m_t) + m_v + m_t}{\alpha\times \gamma},
\end{align*}

\vspace{-0.3em}
where $\alpha$ ratio of the number of accepted tokens among all proposed tokens. In contrast, the average I/O complexity of vanilla decoding, where each token is generated using full attention, is given by:
% $\text{I/O}_{\text{vanilla}} = (n+1) \times (m_v + m_t)$,
% with an average I/O per token of:
$\text{I/O}_{\text{average}}^{\text{vanilla}} = m_v + m_t$. When $\alpha$ is sufficiently large, i.e., $\alpha>(K+m_t)/(m_v+m_t)+\gamma^{-1}$, the average I/O per token in our method becomes considerably lower, resulting in improved decoding efficiency. Intuitively, we hope that the ratio between the numbers of the accepted tokens and all proposed tokens is larger than the ratio between the numbers of retrained KV pairs and the full KV cache. This can be achieved due to the concentration behavior of attention scores in Section~\ref{sec:obs}. The empirical superiority of our method in the next section verifies this inequality in the realistic setting.

% When $k \ll m_v$, and  $\gamma \approx n$, the average I/O per token in our method becomes considerably lower, resulting in improved decoding efficiency.

\section{Experiment}
\label{sec:exp}
\begin{table*}[t]
\centering
\small
\setlength{\tabcolsep}{5pt}
\begin{tabular}{lccccccccc}
\toprule
\multirow{2}{*}{\bf Methods} & \multicolumn{2}{c}{\textbf{MLVU}} & \multicolumn{2}{c}{\textbf{VideoMME-s}} & \multicolumn{2}{c}{\textbf{VideoMME-m}} & \multicolumn{2}{c}{\textbf{VideoMME-l}} \\
 \cmidrule(lr){2-3}\cmidrule(lr){4-5} \cmidrule(lr){6-7} \cmidrule(lr){8-9}
&Acc. \footnotesize{(\%)} &Speedup
&Acc. \footnotesize{(\%)} &Speedup
&Acc. \footnotesize{(\%)}&Speedup
&Acc. \footnotesize{(\%)}&Speedup\\
\midrule
\multicolumn{6}{l}{\textit{\textbf{LLaVA-OneVision-7B}}}\\
LayerSkip&10.0&0.47$\times$&5.6&0.33$\times$&8.1&0.46$\times$&4.8&0.44$\times$\\
Streaming&34.7&1.34$\times$&36.4&1.38$\times$&41.0&1.51$\times$&36.2&1.45$\times$\\
\rowcolor{lightLavender}
\textsc{StD} (ours)&\textbf{47.8}&\textbf{1.72}$\times$&\textbf{51.8}&\textbf{1.82}$\times$&\textbf{52.1}&\textbf{1.83}$\times$&\textbf{52.9}&\textbf{1.59}$\times$ \\
\midrule
\multicolumn{6}{l}{\textit{\textbf{Qwen2-VL-7B-Instruct}}}\\
LayerSkip&5.2&0.63$\times$&3.7&0.59$\times$&4.9&0.55$\times$&5.7&0.55$\times$\\
Streaming&53.9&1.61$\times$&52.9&1.32$\times$&59.2&1.36$\times$&59.6&1.36$\times$\\
\rowcolor{lightLavender}
\textsc{StD} (ours)&\textbf{66.1}&\textbf{1.94}$\times$&\textbf{71.8}&\textbf{1.71}$\times$&\textbf{73.4}&\textbf{1.62}$\times$&\textbf{81.8}&\textbf{1.70}$\times$ \\
\bottomrule
\end{tabular}
\caption{Comparisons of the acceptance rate (Acc.) and wall time speedup of \textsc{StD} and previous draft models. \textbf{Bold} denotes the best method. Since all the methods are lossless, we do not report the evaluation of the generated contents.%\qian{We should also show the performance of each method, and the original one to demontrate that ours is a loseless one?}
% $^\dagger$ denotes significantly better than the best baseline models with $p<0.01$.
}
\label{tbl:main}
\vspace{-1.3em}
\end{table*}

%In this section, we empirically validate that Sparse-to-Dense decoding can accelerate decoding without any loss and uncover several insightful findings.

\vspace{-0.2em}
\paragraph{Baselines.}
To evaluate the effectiveness of our proposed Sparse-to-Dense decoding, we compare it against the following baselines: 
1) Layerskip~\cite{elhoushi2024layer}: This method utilizes a model with an layer-level early exit mechanism to propose draft tokens. %, while a full-layer model verifies the tokens in parallel. 
This baseline is inspired by the work of~\citeauthor{elhoushi2024layer} on text-only LLMs, and originally requires additional training. For a fair comparison with our method, we adapt it to Video-LLMs in a tuning-free manner.
2) Streaming~\cite{chen2024magicdec}: 
This method employs a model with streaming attention~\cite{xiao2023efficient} to propose draft tokens. %, while a full-layer model performs verification in parallel. 
Similar to LayerSkip, this baseline is derived from the work of~\citeauthor{chen2024magicdec} on text-only LLMs. To ensure comparability with our approach, we extend its implementation to Video-LLMs.

\vspace{-0.2em}
\paragraph{Datasets and evaluation metrics.}
We evaluate Sparse-to-Dense on two widely adopted benchmarks: MLVU~\cite{zhou2024mlvu} and VideoMME~\cite{fu2024video}. MLVU is specifically designed for long-duration videos, while VideoMME encompasses short, medium, and long-duration videos, providing a comprehensive assessment across various video lengths.
For our evaluation, we adhere to the protocols established in previous works on speculative decoding. We report two primary metrics: \textbf{\textit{acceptance rate}} of the draft tokens and wall time \textbf{\textit{speedup}}.

%Acceptance accuracy quantifies the probability that the original model accepts a token proposed by the fast and less accurate model. Speedup measures the computational efficiency gained through our approach relative to the original model.
%It is important to note that we do not report answer accuracy in our evaluation. Both our Sparse-to-Dense method and the baseline models are lossless acceleration techniques, meaning they maintain identical performance levels to the original model.

\vspace{-0.2em}
\paragraph{Implementation Details.}
Our experiments are conducted using widely adopted state-of-the-art Video-LLMs, specifically LLaVA-OneVision (7B)~\cite{li2024llava} and Qwen2-VL (7B)~\cite{wang2024qwen2}. 
%Qwen2-VL employs a ViT-based vision encoder derived from DFN, while LLaVA-OneVision utilizes the SigLIP vision encoder.
We prompt the Video-LLMs to generate chain-of-thought~\cite{wei2022chain} responses to enhance their performance.
We set the sum of the textual token count $m_t$ and the selected visual KV cache count $K$ to 1024, with a batch size of 8. 
The number of tokens verified by the dense model $\mathcal{M}_d$ is fixed at $\gamma=9$. 
The ablation of hyperparameters can be found in Appendix Section~\ref{sec:abl}.
Our framework is implemented based on Hugging Face's Transformers library. 
All experiments are conducted on NVIDIA A100 GPUs with 80 GB of memory, and are repeated three times with
different random seeds, and the average results are reported.

\vspace{-0.2em}
\paragraph{Main Results}
Table~\ref{tbl:main} summarizes the performance across various reasoning tasks. We have the following findings:
1) \textit{The draft model based on LayerSkip performs worse than that utilizing sparse attention} (e.g., Streaming and \textsc{StD}). The primary reason for this discrepancy is that LayerSkip causes a substantial distributional shift between the draft model and the target model, leading to a low acceptance rate. Although the draft model with layer skipping runs considerably faster than the sparse attention counterparts, this advantage is insufficient to compensate for the overall wall-time speedup loss introduced by layer skipping.
2) \textit{Draft models based on sparse attention generally provide more wall time speedup.} Whether in \textsc{StD} or Streaming, we observe a consistently high acceptance rate. This indicates that, for most of the time, the target model does not require the full KV cache but only a sparsely selected subset cache. However, it is important to note that since LLMs perform autoregressive decoding, an incorrect token can propagate errors to subsequent tokens. Thus verification with the full KV cache is essential.
3) \textit{Our model outperforms the streaming-based draft model}, achieving 62.2\% in acceptance length and 1.74$\times$ in wall-time speedup on average. This advantage stems from our method’s ability to leverage the unique characteristics of Video-LLMs to select important KV cache. As observed in section~\ref{sec:obs}, text-guided video cache selection effectively identifies and retains the most critical cache elements.

\vspace{-0.7em}
\section{Conclusion}
\label{sec:con}
\vspace{-0.7em}
We introduce \textsc{StD}, a training-free, plug-and-play decoding method that employs sparse top-$K$ attention as the draft model in speculative decoding while leveraging full attention for verification in parallel, ensuring lossless acceleration. Extensive experiments demonstrate that \textsc{StD} significantly outperforms strong baselines that use LayerSkip and Streaming as the draft models. Overall, \textsc{StD} achieves up to a 1.94$\times$ walltime speedup while maintaining identical output quality.
In the future, we hope to extend our work to accelerate long CoT Video-LLMs such as QvQ~\cite{qvq72b_preview}.
\section*{Limitation}
A notable limitation of our current approach is that all KV caches are still stored in GPU memory (i.e., HBM). While HBM provides the high bandwidth necessary for fast computations, its capacity is inherently limited, which poses a significant bottleneck during inference—especially as model sizes and sequence lengths increase. The limited HBM capacity may lead to restricted batch size.

In the future, a promising solution to this challenge is to offload portions of the KV caches to CPU memory. Although CPU memory typically has lower bandwidth compared to HBM, it offers substantially larger capacity. By developing efficient data transfer and caching strategies, it may be possible to mitigate the HBM bottleneck without sacrificing inference accuracy, thereby enabling more scalable and efficient processing for large Video-LLMs.
\bibliography{custom}
\clearpage
\appendix
\section{Preliminary}
\label{sec:pre}
\paragraph{Speculative Decoding}
% Before introducing Sparse-to-Dense decoding,
We first formalize our notation and provide a brief overview of the speculative decoding in autoregressive LLMs, which is the key background knowledge for our method.
We represent the input sequence for a Video-LLM as a combination of visual tokens and textual tokens. 
Specifically, the visual tokens are denoted as $X_v = \{x_0, \cdots, x_{m_v-1}\}$, and the textual prompt is denoted as $X_t = \{x_{m_v}, \cdots, x_{m-1}\}$. 
Here, $m_v$ is the number of visual tokens, $m_t$ is the number of textual tokens, and the total input sequence length is $m=m_v+m_t$.
% The total number of tokens, including both the input and the generated responses, is denoted as $n$.
The key and value cache for token $x_i$ are represented by $K_{x_i}$ and $V_{x_i}$, respectively.

\paragraph{Inference of Auto-regressive Models.}
The inference stage of auto-regressive models, e.g., Video-LLMs, can be divided into two stages: 1) \textit{prefilling}: 
The video LLM processes the input sequence, which includes both visual tokens 
$X_v$ and textual tokens $X_t$, in an autoregressive and parallel manner.
For each token $x_i$ in the combined input $\{X_v, X_t\}$, the model computes and stores the corresponding KV cache entries.
This stage effectively encodes the input sequence and prepares the model for generating a response.
The output of this stage is the first token $x_{m}$  of the model's response.
2)~\textit{decoding}:
After prefilling, the model enters the decoding phase, generating output tokens sequentially.
At each decoding step $j = m+1, m+2, \cdots$, the video LLM generates a new token $x_{j}$ based on the KV cache from all prior tokens.
After generating, the KV cache is updated with each newly generated token.
This process continues iteratively until a stopping criterion is met, such as reaching an end-of-sequence token or hitting a maximum token limit.

\section{Related Works}
\label{sec:related}

\paragraph{Sparse Attention in MLLMs}
Normally, an image or a video frame is represented as a large number of tokens in MLLMs, e.g., $196$ visual tokens per image in VILA~\cite{lin2024vila}, which significantly impacts the computational and storage during model training and inference. 
Visual token compression aims to reduce the number of visual tokens to address it directly. 
% Some works mainly focus on efficiently reducing visual tokens for high-resolution images, for instance, by using diverse resolutions~\cite{dong2024internlm} or simultaneously extracting low-resolution and high-resolution information~\cite{huang2024hires,hu2024mplug}.
The majority of visual token compression methods either train from scratch or perform additional training based on existing models.
For example, some image-based MLLMs rely on vision-language alignment~\cite{cao2024madtp,yao2024deco,song2024less} or aggressively removing all visual tokens after a certain layer~\cite{wen2024efficient}, while methods designed for video-based MLLMs consider the unique characteristics of video, such as employing memory mechanisms~\cite{lan2024vidcompress} or compressing tokens along spatial and temporal dimensions sequentially~\cite{shen2024longvu}.
A smaller portion of works study the test-time (training free) visual token compression for accelerating the inference procedure. FastV~\cite{chen2024image} performs pruning by analyzing the attention pattern from shallow layers and deep layers, while another approach directly applies full visual token removal during the inference stage~\cite{lin2024boosting}.
In our method, \textsc{StD}, the design of the drafter model is related to training-free visual token compression techniques. However, these previous methods inevitably impact the original model's performance. In contrast, we propose to utilize visual token compression as a drafter model to achieve lossless inference acceleration.

\paragraph{Speculative Decoding} Speculative decoding is proposed by \cite{leviathan2023fast} and \cite{chen2023accelerating} to accelerate the inference of LLMs, where the throughput of LLMs is improved $2\sim 3$ times without sacrificing the performance. The algorithm consists of two stages: drafting and verification. The drafting stage adopts a small model (drafter) to generate a long sequence of possible future tokens swiftly, while the verification stage accepts a part of the tokens predicted in the drafting stage in a token-by-tone manner. The follow-up improves the speculative decoding from these two perspectives. Specinfer~\citep{miao2024specinfer}, Eagle~\citep{li2024eagle} and Medusa~\citep{cai2024medusa} propose to train a drafter to generate tokens with a tree structure, and the verification is conducted on the tree in a branch-by-branch manner. Hu and Huang~\citep{huaccelerated} also organize the draft tokens as a tree, but they verify the tokens in a branch as a whole. Glide~\citep{du2024glide} generates draft tokens as an unbalanced tree, which alleviates the burden of the drafter while achieving significant acceleration. SpecTr~\citep{sun2024spectr} views speculative decoding from the optimal transport view and proposes to verify a batch of draft tokens jointly. They show that the proposed algorithm is optimal up to a multiplicative factor. Sun et al.~\citep{sun2024optimal} boot the acceleration by a joint verification of a single draft trajectory. Instead of using a token-by-token manner, they accept the draft sentences as a whole. Lie et al.~\citep{liu2023online} proposes to update the parameters of drafters in an online manner, which is shown to be effective in various applications. MagicDec~\citep{chen2024magicdec} analyzes the speculative decoding in the long-context setting with an emphasis on the FLOPS and memory. SpecExec~\citep{svirschevski2024specexec} focuses on a special setting where the LLMs are offloading their parameters. Several works~\citep{gagrani2024speculative,jang2024lantern,teng2024accelerating} study the speculative decoding of MLLMs. However, they focus either on the image understanding problem or the image generation problem. In contrast, our work is the first to study video generation acceleration via speculative decoding.

\section{Ablation Stuidies}
\label{sec:abl}
\begin{figure}
\centering
\vspace{-.2cm}
\subfloat[]{
\includegraphics[width=0.48\textwidth]{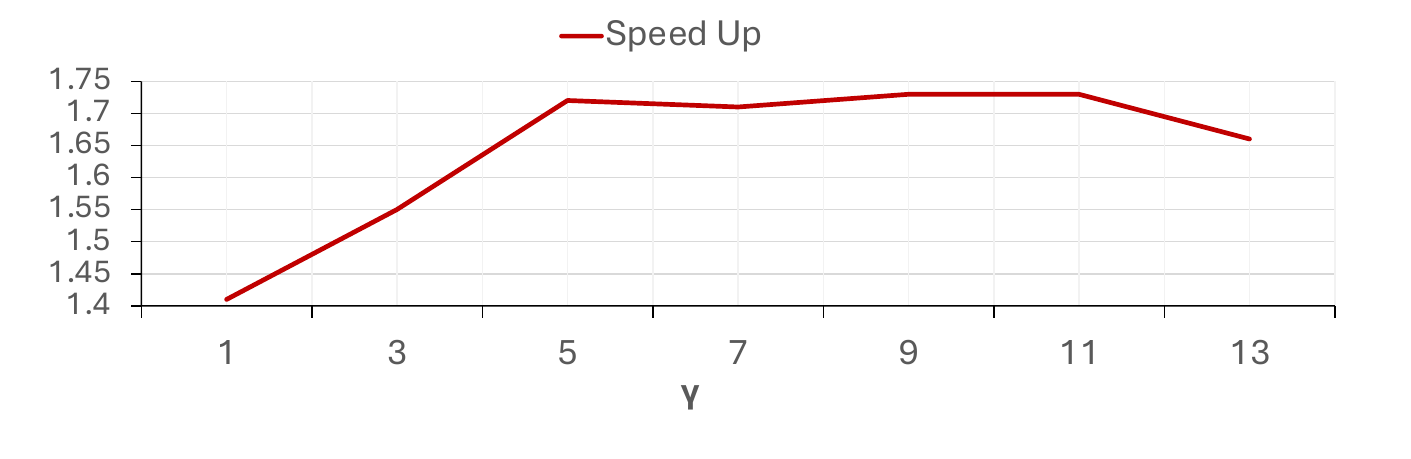}
\label{fig:abla}
}
\hspace{0.1cm}
\subfloat[]{
\includegraphics[width=0.48\textwidth]{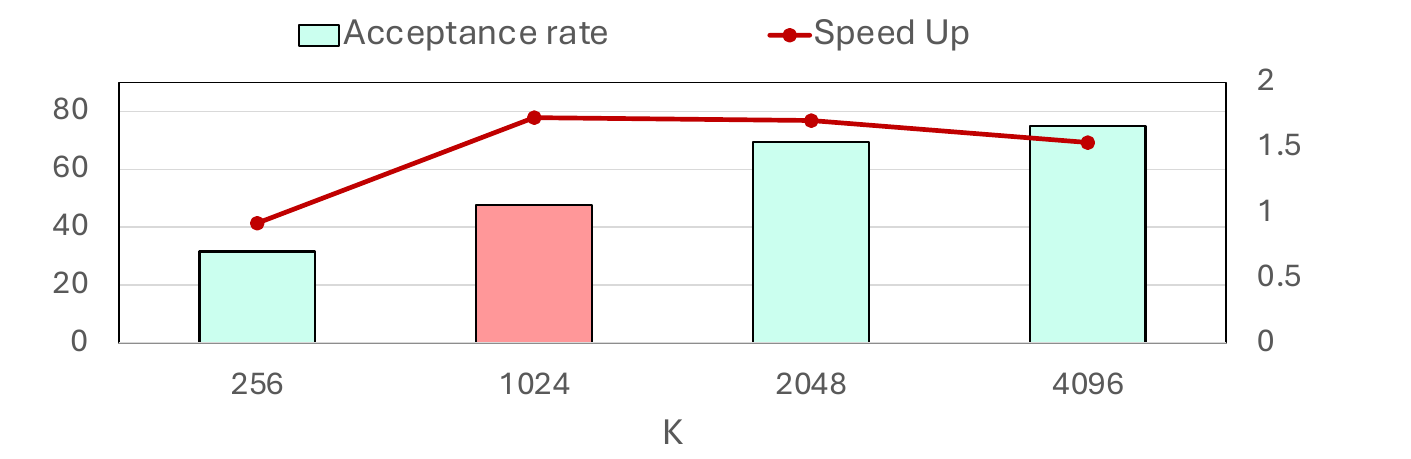}
\label{fig:ablb}
}
\caption{
 Effect of $K$ and $\gamma$ on MLVU using LLaVA-OneVision-7B.
}
\label{fig:app_example}
\end{figure}

We also conducted additional experiments to analyze the impact of hyperparameters ($\gamma$ and $K$) on model performance.
As shown in Figure~\ref{fig:abla}, we can see that as $gamma$ increases, the speed up gradually improves. This improvement is because the sparse model makes accurate predictions, which allows the computational overhead to be spread out over more tokens. However, when $gamma$ reaches 13, the speed up starts to decline because the model's accuracy in correctly predicting 13 consecutive tokens is insufficient.
At the same time, as shown in Figure~\ref{fig:ablb}, when $K$ is small, the acceptance rate is low, resulting in a lower speed up. In contrast, when $K$ is large, the sparse model is not as fast, which also leads to a reduced speed-up.
% \section{Example Appendix}
% \label{sec:appendix}

% This is an appendix.

% video 的大kv cache，面临challenge
% 针对这个问题， 有一些general的思路，比如kv compression， 比如**有损**/ 无损 
% inspired by speculative decoding。
% 我们方法的做法， 实验中的结果， contribution。

\end{document}